\documentclass{article}
\usepackage{spconf,amsmath,graphicx}
\graphicspath{{figures/}}
\usepackage{algorithm}
\usepackage{algorithmic}
\usepackage{booktabs}
\usepackage{float}
\usepackage{xcolor}

 \usepackage{graphicx}

\title{A Curriculum Learning Approach for Multi-domain Text Classification Using Keyword weight Ranking}
\makeatletter
\def\name#1{\gdef\@name{#1\\}}
\makeatother
\name{\em{Zilin Yuan$^{1}$,
    Yinghui Li$^{1}$,
    Yangning Li$^{1}$,} 
    Rui Xie$^{2}$, 
    Wei Wu$^{2}$,
    Hai-Tao Zheng$^{1,3*}$
\thanks{* Corresponding author. (E-mail: zheng.haitao@sz.tsinghua.edu.cn)}
}

\address{$^{1}$Shenzhen International Graduate School, Tsinghua University \\
      $^{2}$Meituan, $^{3}$Peng Cheng Laboratory}
\begin{document}
\ninept
\maketitle
\begin{abstract}
Text classification is a very classic NLP task, but it has two prominent shortcomings: On the one hand, text classification is deeply domain-dependent. That is, a classifier trained on the corpus of one domain may not perform so well in another domain. On the other hand, text classification models require a lot of annotated data for training. However, for some domains, there may not exist enough annotated data. Therefore, it is valuable to investigate how to efficiently utilize text data from different domains to improve the performance of models in various domains. Some multi-domain text classification models are trained by adversarial training to extract shared features among all domains and the specific features of each domain. We noted that the distinctness of the domain-specific features is different, so in this paper, we propose to use a curriculum learning strategy based on keyword weight ranking to improve the performance of multi-domain text classification models. The experimental results on the Amazon review and FDU-MTL datasets show that our curriculum learning strategy effectively improves the performance of multi-domain text classification models based on adversarial learning and outperforms state-of-the-art methods.
\end{abstract}
\section{Introduction}
\label{sec:intro}
Text classification is one of the fundamental NLP tasks and it has a wide range of applications, such as spam determination~\cite{sahmoud2022spam}, news classification~\cite{nugroho2021large}, and evaluation of e-commerce products~\cite{agarap2018statistical}. The research on text classification methods can be traced back to the methods based on expert rules in the 1950s. In the 1990s, machine learning classification methods combining feature engineering and classifiers began to appear~\cite {McCallum1998ACO}, and now the more popular method is to use CNN~\cite{zhang2015sensitivity}, RNN\cite{liu2016recurrent,lai2015recurrent}, attention mechanism~\cite{yang2016hierarchical} and other deep learning methods for classification.

\begin{figure}[]
    \centering
    \includegraphics[width=8.5cm]{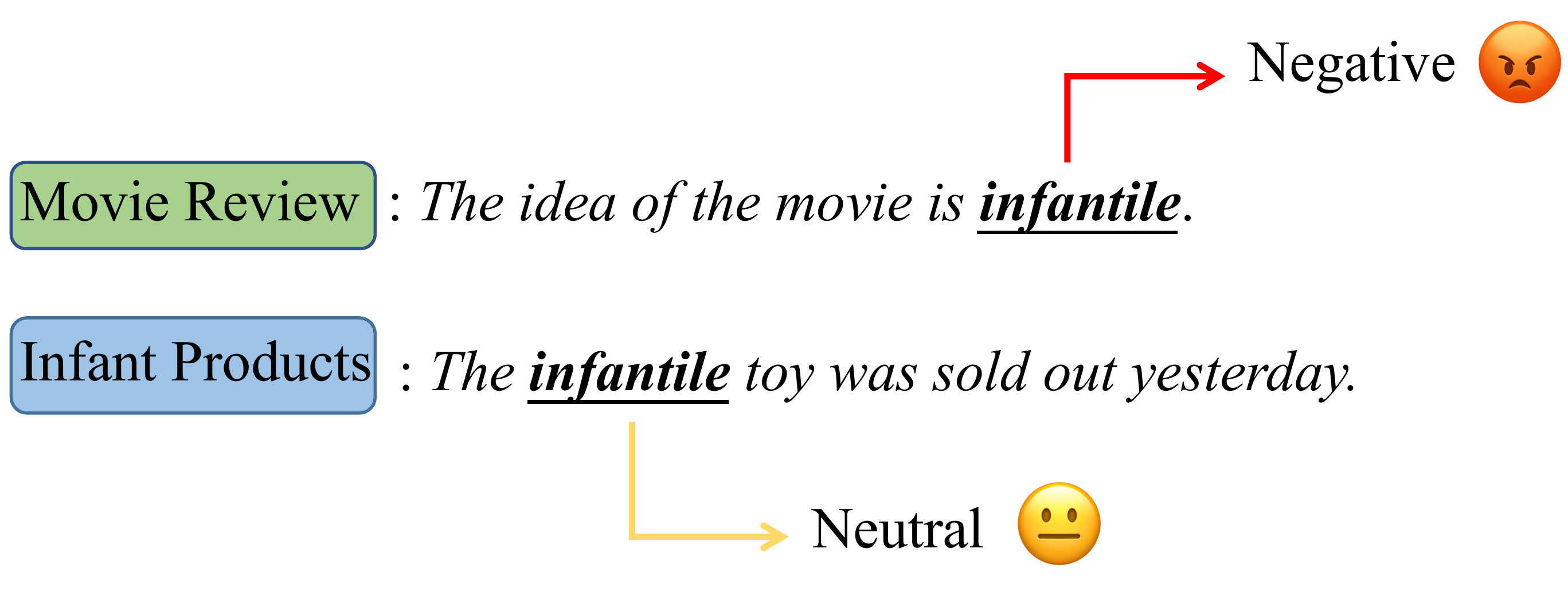}
    \caption{The different sentiments of “infantile” in different domains.}
    \label{fig:intro_example}
\end{figure}

But no matter which method, there are two main problems: the highly domain-dependence and the need for amounts of the annotated corpus. Domain-dependence means that the classifier trained on a certain domain may not have the same effect in other domains, because the meaning of vocabulary of different domains may be different, and even the same word expresses different meanings in different domains. As shown in Figure~\ref{fig:intro_example}, the “infantile”~\cite{cai2019multi} often expresses a negative meaning in the domain of Movie Review (e.g., “The idea of the movie is infantile”), but there is usually no obvious emotional color in the evaluation of Infant Products (e.g., “The infantile toy was sold out yesterday”). Therefore, when we want to train classifiers in different domain texts, we need enough labeled data in each domain, but not all domains have enough domain corpus to train. So it is necessary to make full use of the corpus in different domains to classify the texts in a specific domain, also known as the Multi-Domain Text Classification (MDTC)~\cite{li2008multi,chen2018multinomial}. However, the traditional MDTC methods~\cite{chen2018multinomial,luo2022mere} all ignore a piece of important information. That is, the classification difficulty of different domains is different.

\indent The difficulty of text classification of different domains is inconsistent, so this feature might be used to make the model learn the data from easy to difficult. This way of learning is like human learning, in which simple lessons are learned first, followed by complex lessons. This learning mode is called curriculum learning~\cite{bengio2009curriculum}, and it has shown outstanding promotion in NLP tasks such as dialog state tracking~\cite{dai2021preview}, few-shot text classification~\cite{wei2021few}, Chinese Spell Checking~\cite{DBLP:journals/corr/abs-2207-09217} and so on. The core of the course learning lies in the difficulty measurer of data samples and the data scheduler. Combined with the extraction of private and shared features by multi-domain text classification, we propose that the sum of the weights of domain keywords can be regarded as a measurer of the difficulty of domain-specific feature extraction to adjust the order when the corpus of a specific domain is fed into the model.

Based on the above motivations, we propose a framework called Keyword-weight-aware Curriculum Learning (KCL) for MDTC, which includes the following two features:

\indent 
1) By calculating the word weights of texts, take the Top-N words as the domain keywords, and calculate the sum of the weights of these N keywords to measure the difficulty of extracting the domain-specific feature of each domain. The higher the sum is, the more obvious the domain-specific features are, and the easier it is to extract, so it is necessary to enter the model for training earlier.

\indent 
2) Using different methods of keyword extraction and testing different numbers of keywords to find the best order of domains.

\indent 
The experimental results show that our proposed approach improves MDTC performance and achieves new state-of-the-art results on the Amazon review dataset and FDU-MTL dataset.

\section{METHOD}
\label{sec:format}
\begin{figure}[htp]
    \centering
    \includegraphics[width=8.5cm]{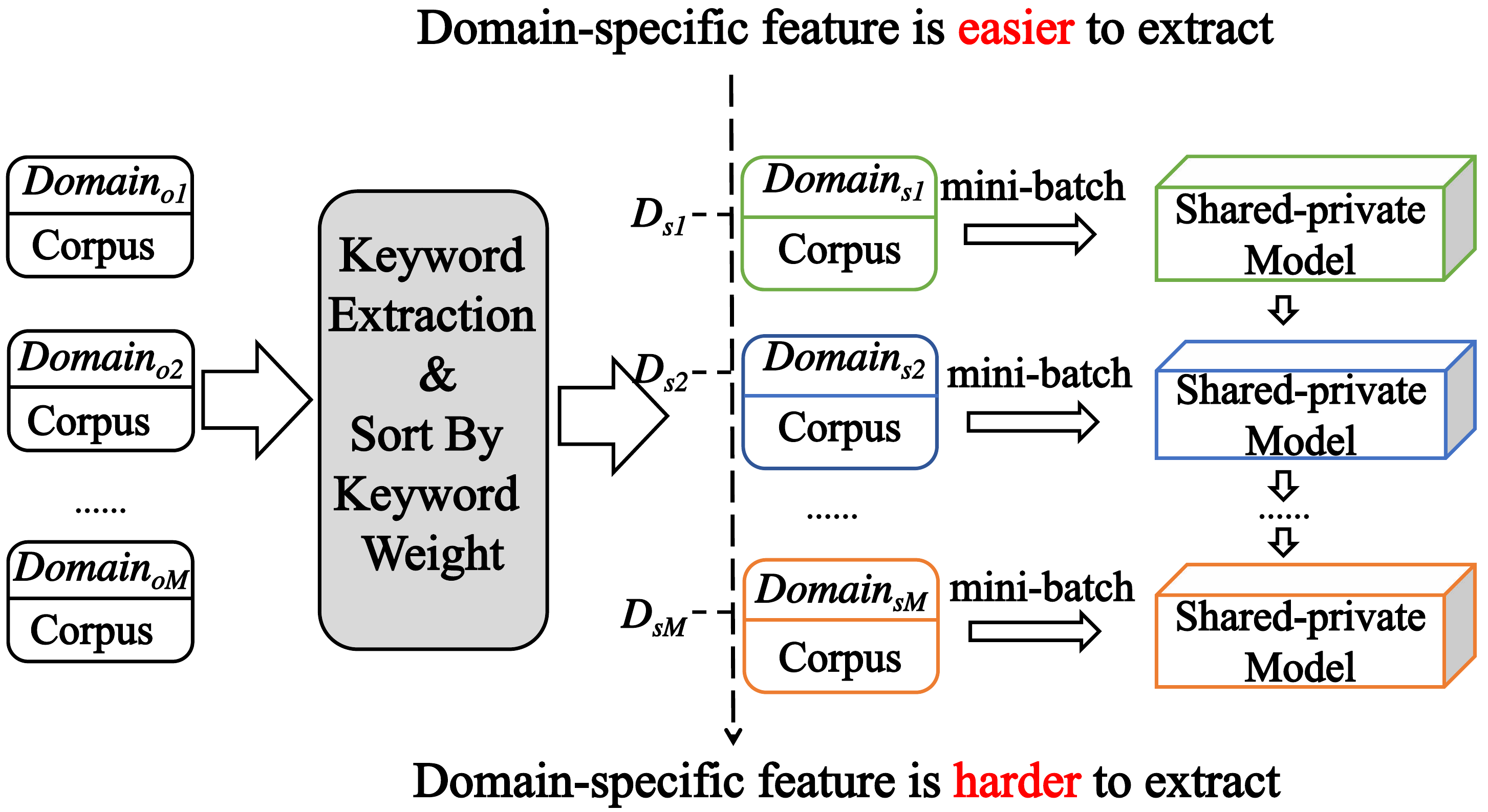}
    \caption{The architecture of KCL.}
    \label{fig:KCL}
\end{figure}
\noindent As shown in Figure~\ref{fig:KCL}, KCL includes two parts: One part contains keyword extraction and sorting the domains in order of the weight of keywords, and another is the shared-private model~\cite{chen2018multinomial}.

\subsection{Keyword Extraction and Summarization}
\begin{figure}[htp]
    \centering
    \includegraphics[width=8.5cm]{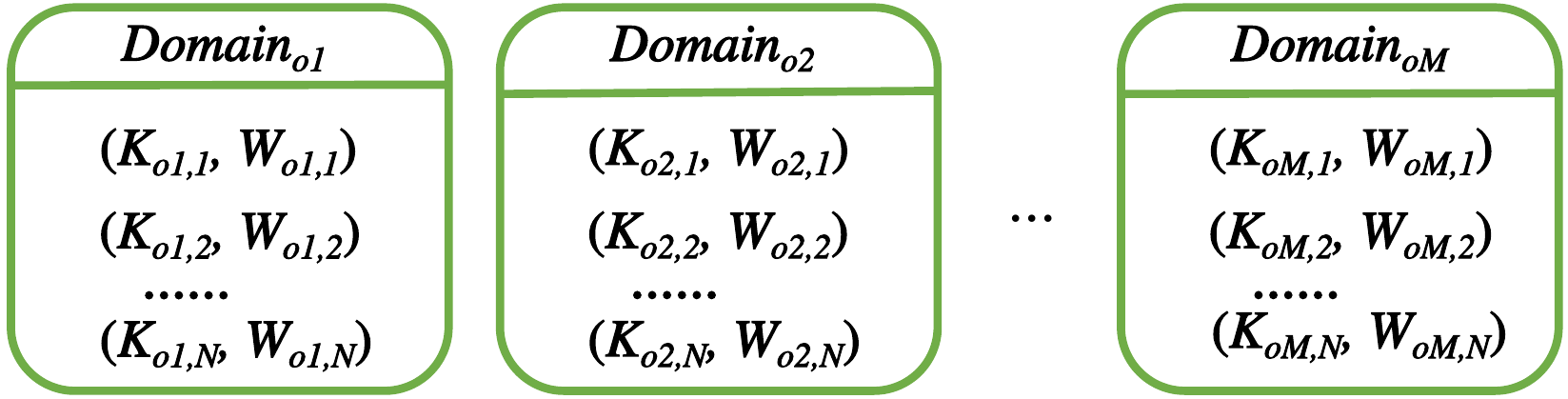}
    \caption{The sorted words list and its weight of each domain. $K_{oi,j}$ means the $j^{th}$ keyword of the $Domain_{oi}$, and the $W_{oi,j}$ means the weight of the keyword.}
    \label{fig:wordsList}
\end{figure}

The core of KCL is to first extract the Top-N keywords of each domain through the keyword extraction algorithm and then measure the apparent degree of domain-specific features in each domain by calculating the sum of the weights of these N keywords. There are M original domains: $Domain_{o1}$, $Domain_{o2}$,...,$Domain_{oM}$, and then we apply the keyword extraction algorithm to the corpus of these M domains, calculate the weights of all words in each domain corpus, and regard the N words with the highest weight of each domain as the keywords of each domain. The keywords obtained from each domain are shown in Figure~\ref{fig:wordsList}.

\indent Then the obviousness of domain-specific feature of $Domain_{oi}$ is calculated as Equation~\ref{eq:woi}:
\begin{equation}
    W_{oi}=W_{oi,1}+W_{oi,2}+...+W_{oi,N}.\label{eq:woi}
\end{equation}    
The larger the $W_{oi}$ is, the more obvious the domain-specific features of the domain are, and the easier it is to be extracted, so the order of entering the model in each step is earlier. After sorting, we can get a sorted domain list: [$Domain_{s1}, Domain_{s2},..., Domain_{sM}$], which is sorted according to the $W$ from high to low. We abbreviate the domain list as [$D_{s1}, D_{s2}, ..., D_{sM}$] in Figure~\ref{fig:KCL}.

\subsection{Shared-private Model}
\begin{figure}[htp]
    \centering
    \includegraphics[width=8.5cm]{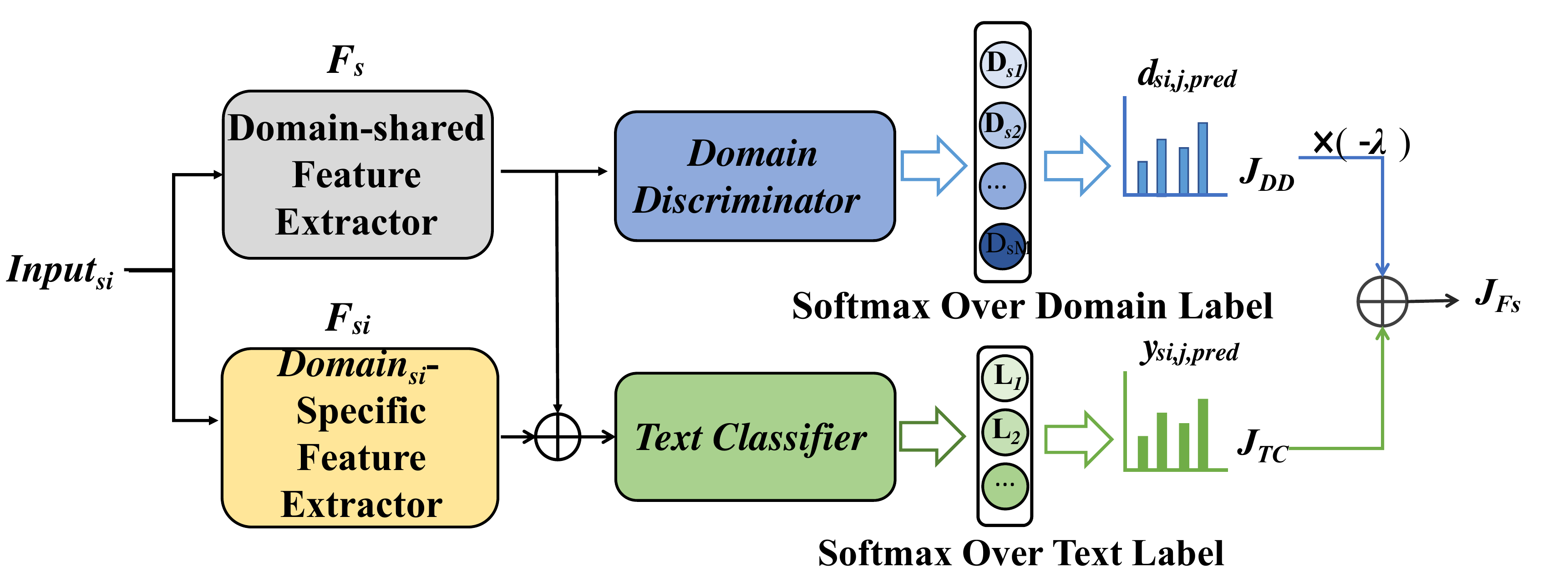}
    \caption{The architecture of the shared-private model.}
    \label{fig:shared-private}
\end{figure}

Getting the sorted domain list, we then sample mini-batch from each domain in the order of [$D_{s1}, D_{s2},..., D_{sM}$] and input them into the shared-private model in every training step. Following ~\cite{chen2018multinomial}, the structure of the shared-private model of KCL is shown in Figure~\ref{fig:shared-private}. $Input_{si}$, as a mini-batch from $D_{si}$, would enter the domain-shared feature extractor and the $Domain_{si}$-specific feature extractor of $D_{si}$ respectively. After the two extractors finish processing the mini-batch, the output of the domain-shared feature extractor enters the $Domain Discriminator$ and is concatenated with the output of the $Domain_{si}$-specific feature extractor and enters the $Text Classifier$. The $Domain Discriminator$ is used to judge the source domain of the sample. After softmax, we can obtain the probability of the sample from each domain [$D_{s1}, D_{s2}$, ..., $D_{sM}$]. $J_{DD}$ is the objective function of the $Domain Discriminator$ and it can be calculated as Equation~\ref{eq:jdd}. We denote $sample_{si,j}$=($x_{si,j}$,$y_{si,j}$,$D_{si}$) as the $j^{th}$ sample of $D_{si}$, in which $x_{si,j}$ means the text and $y_{si,j}$ means the text label.

\begin{table*}[ht]
\small
    \centering
    \caption{The accuracy of different keyword extraction methods on FDU-MTL dataset.}
    \begin{tabular}{*{10}{c}}
    \toprule
        \textbf{domain} &  \textbf{CNN} & \textbf{BERT} & \textbf{CAN} & \textbf{CRAL} & \textbf{COBE} & \textbf{KCL-random} & \textbf{KCL-YAKE} & \textbf{KCL-TextRank}& \textbf{KCL-KeyBERT}\\ 
    \midrule
        Books       &85.30 &87.00 &87.80 &89.30 &90.17 &92.42 &93.00 &\bf93.42 &93.08\\ 
        Electronics &87.80 &88.30 &91.60 &89.10 &93.58 &93.5 &93.33 &94.00 &\bf94.92\\
        DVD         &76.30 &85.60 &89.50 &{\bf91.00} &89.67 &88.42 &89.42 &89.58 &89.92\\
        Kitchen     &84.50 &91.00 &90.80 &92.30 &91.50 &91.08 &92.08 &{\bf92.67} &92.50\\
        Apparel     &86.30 &90.00 &87.00 &91.60 &92.33 &92.25 &92.42 &92.08 &{\bf92.67}\\
        Camera      &89.00 &90.00 &93.50 &96.30 &93.58 &91.92 &93.00 &{\bf93.92} &93.67\\
        Health      &87.50 &88.30 &90.40 &87.80 &93.92 &94.42 &94.33 &94.75 &{\bf95.67}\\
        Music       &81.50 &86.80 &86.90 &88.10 &90.33 &89.08 &88.50 &\bf91.00 &90.42\\
        Toys        &87.00 &91.30 &90.00 &91.60 &93.42 &92.5 &92.75 &\bf93.67 &93.33\\
        Video       &82.30 &88.00 &88.80 &{\bf92.60} &89.91 &88.58 &91.00 &90.67 &91.67\\
        Baby        &82.50 &91.5 &92.00 &90.90 &93.92 &93.58 &93.50 &\bf94.75 &94.58\\
        Magazine    &86.80 &92.8 &94.50 &95.20 &94.08 &91.67 &92.67 &93.50 &\bf94.17\\
        Software    &87.50 &89.3 &90.90 &87.70 &93.42 &93.75 &92.33 &\bf95.00 &94.33\\
        Sports      &85.30 &90.8 &91.20 &91.30 &92.83 &92.67 &93.41 &94.25 &\bf94.42\\
        IMDB        &83.30 &85.80 &88.50 &90.80 &86.91 &89.08 &89.83 &90.42 &\bf90.83\\
        MR          &79.00 &79.00 &77.10 &77.30 &84.33 &82.33 &84.75 &84.33 &\bf85.58\\
    \midrule
        Avg         &84.30 &84.30 &89.40 &90.20 &91.49 &91.07 &91.64 &92.33 &{\bf 92.62}\\
    \bottomrule
    \end{tabular}
    \label{table:keyword_methos_FDU}
\end{table*}

\begin{table*}[htbp]
\small
    \centering
    \caption{The accuracy of different keyword extraction methods on Amazon review dataset.}
    \begin{tabular}{*{11}{c}}
    \toprule
        \textbf{domain} &  \textbf{MLP} & \textbf{MAN} & \textbf{CAN} & \textbf{CRAL} & \textbf{KCL-random} &  \textbf{KCL-YAKE} & \textbf{KCL-TextRank}& \textbf{KCL-KeyBERT}\\ 
    \midrule
        Books        &82.40 &82.98 &83.76 &85.26 &89.25 &91.00 &\bf91.83 &89.75\\ 
        Electronics  &82.15 &84.03 &84.68 &85.83 &90.50 &89.50 &\bf91.75 &91.67\\
        DVD          &85.90 &87.06 &88.34 &89.32 &90.33 &90.25 &\bf91.33 &91.08\\
        Kitchen      &88.20 &88.57 &90.03 &91.60 &91.08 &91.33 &\bf93.42 &92.58\\
       
    \midrule
        Avg         &84.66 &85.66  &86.70 &88.00 &90.29 &90.51 &\bf92.08 &91.27\\
    \bottomrule
    \end{tabular}
    \label{table:keyword_methos_amazon}
\end{table*}

\begin{equation}
    J_{DD}=-\sum_{i=1}^{M}\sum_{sample_{si,j} \in D_{si}} logP(d_{{si,j,pred}}=D_{si}).  \label{eq:jdd}
\end{equation} 

$d_{si,j,pred}$ represents the domain prediction of the $j^{th}$ sample in the $D_{si}$, and it is calculated as Equation~\ref{eq:dijpred}. $P(d_{{si,j,pred}}=D_{si})$ means the probability that the prediction of the domain is right.
\begin{gather}
    d_{si,j,dist}=DomainDiscriminator(F_s(x_{si,j})), \\
    d_{si,j,pred}=softmax(d_{si,j,dist}).  \label{eq:dijpred}
\end{gather} 

The $Text Classifier$ is used to classify the sample, and the output is the logits of the different labels [$L_{1}$, $L_{2}$, ...]. After softmax, we can obtain the probability that the predicted label is equal to the true label. The objective function $J_{TC}$ of the $Text Classifier$ can be calculated as Equation~\ref{eq:JTC}:
\begin{equation}
    J_{TC}=-\sum_{i=1}^{M}\sum_{sample_{si,j} \in D_{si}}logP(y_{si,j,pred}=y_{si,j}). \label{eq:JTC}
\end{equation}  

$y_{si,j,pred}$ represents the prediction of the $j^{th}$ sample in the $D_{si}$, and it is calculated as Equation \ref{eq:yijpred}. $P(y_{si,j,pred}=y_{si,j})$ means the probability that the prediction of the text label is right.
\begin{gather}
    y_{si,j,dist}=TextClassifier(concate(F_s(x_{ij}),F_{si}(x_{ij}))),\\
    y_{si,j,pred}=softmax(y_{si,j,dist}), \label{eq:yijpred}
\end{gather}  
where $concate(F_s(x_{ij}),F_{si}(x_{ij}))$ means concatenating the output of $F_s$ and $F_{si}$.

For each domain-specific feature extractor, it aims to help the $TextClassifier$ to better classify the samples, so its objective function is similar to $J_{TC}$. The objective function of the $F_{si}$ is calculated as Equation~\ref{eq:jfdi}:
\begin{equation}
    J_{F_{si}}=-\sum_{sample_{si,j} \in D_{si}} logP(y_{si,j,pred}=y_{si,j}).  \label{eq:jfdi}
\end{equation}  

For the domain-shared feature extractor $F_s$, it not only makes  $TextClassifier$ more accurate but also interferes with the judgment of $DomainDiscriminator$ as much as possible. Because if the text from different domains passes through the shared feature extractor, and $Domain Discriminator$ cannot identify the source domain of the sample, it means that the shared features extracted by the $F_s$ are exactly domain invariant. So its objective function needs to combine the loss of $Text Classifier$ with the loss of $DomainDiscriminator$. The objective function of the $F_s$ is as Equation~\ref{eq:jfs}:
\begin{equation}
    J_{F_s}=J_{TC}+J_{DD}\cdot(-\lambda) . \label{eq:jfs}
\end{equation}
$\lambda$ is a hyperparameter larger than 0 to balance the weight of $TextClassifier$ and $DomainDiscriminator$ in the shared feature extractor. Since the domain-shared feature extractor aims to interfere with the judgment of $DomainDiscriminator$, $\lambda$ is multiplied by minus one.

\section{EXPERIMENTS}
\label{sec:pagestyle}

\subsection{Datasets}
We use two classic datasets in multi-domain text classification domain for experiments: Amazon review dataset~\cite{fang2014domain,dong2021survey} and FDU-MTL dataset~\cite{liu2017adversarial}. Amazon review dataset contains product reviews in 4 domains: DVD, Books, Electronics, and Kitchen. There are 1000 positive and negative reviews in each domain. We divide these data into training set and test set according to the ratio of 4:1, and mix the positive and negative reviews in random order as the total training set and test set of each domain.

The FDU-MTL dataset is much larger than the Amazon review dataset, with a total of 16 domains: Books, Electronics, DVD, Kitchen, Apparel, Camera, Health, Music, Toys, Video, Baby, Magazine, Software, Sports, IMDB, and MR. The first 14 of them are product reviews from amazon, and IMDB and MR are movie reviews crawled from IMDB and Rotten Tomatoes, respectively. Each domain contains around 1600 labeled samples in training set, 400 labeled samples in test set, and 2000 unlabeled samples. These unlabeled sample data fit to train the $Domain Discriminator$.

\subsection{Discussion: the method of extracting keyword}
Different keyword extraction methods will bring different effects when extracting and calculating keyword weights. That is, different keywords may be generated, and the weights of keywords may also be different. Therefore, in this part, selecting 50 keywords, we design experiments to observe the effects of different keyword extraction methods: YAKE, TextRank, and KeyBERT.

\begin{table}[ht]
    \centering
    \small
    \caption{The accuracy of different numbers of keywords on FDU-MTL dataset, using KeyBERT as the keyword selection method.}
    \begin{tabular}{*{10}{c}}
    \toprule
        \textbf{domain} &  \textbf{30} & \textbf{40} & \textbf{50} & \textbf{60} & \textbf{70} & \\ 
    \midrule
        Books       &92.67&91.58&\bf93.08&91.50&91.25\\ 
        Electronics &94.25&93.83&\bf94.92&94.33&94.42\\
        DVD         &89.25&87.75&\bf89.92&88.58&89.00\\
        Kitchen     &92.42&92.17&\bf92.50&91.67&91.67\\
        Apparel     &92.08&91.42&\bf92.67&91.83&92.42\\
        Camera      &92.75&92.83&\bf93.67&92.83&92.08\\
        Health      &94.33&93.75&\bf95.67&93.92&93.75\\
        Music       &89.33&88.08&\bf90.42&89.5&88.83\\
        Toys        &92.92&92.92&\bf93.33&92.67&92.33\\
        Video       &89.00&88.50&\bf91.67&89.00&89.83\\
        Baby        &93.67&93.17&\bf94.58&93.42&93.17\\
        Magazine    &92.33&91.58&\bf94.17&92.42&91.67\\
        Software    &93.25&92.25&\bf94.33&93.25&92.83\\
        Sports      &92.67&93.08&\bf94.42&93.33&93.00\\
        IMDB        &89.75 &90.00 &90.83 &90.33 &\bf91.42\\
        MR          &82.58 &84.25 &\bf85.58 &82.08 &83.25\\
    \midrule
        Avg         &91.46 &91.06 &\bf92.62 &91.29 &91.21\\
    \bottomrule
    \end{tabular}
    \label{table:keyword_number_FDU}
\end{table}

\begin{table}[htbp]
    \centering
    \small
    \caption{The accuracy of different numbers of keywords on Amazon review dataset, using TextRank as the keyword selection method.}
    \begin{tabular}{*{10}{c}}
    \toprule
        \textbf{domain} &  \textbf{30} & \textbf{40} & \textbf{50} & \textbf{60} & \textbf{70} & \\ 
    \midrule
        Books       &88.67&\bf92.67&91.83&89.17&89.89\\ 
        Electronics &89.83&90.92&\bf91.75&89.67&88.83\\
        DVD         &88.92&\bf91.58&91.33&89.42&89.33\\
        Kitchen     &92.17&92.83&{\bf93.42}&90.92&91.58\\
        
    \midrule
        Avg         &89.90&92.00&{\bf92.08}&89.79&89.89\\
    \bottomrule
    \end{tabular}
    \label{table:keyword_number_amazon}
\end{table}

The results of our proposed model and other baselines (including CAN~\cite{wu2021conditional},CRAL~\cite{wu2022co},COBE~\cite{luo2022mere},MLP\cite{chen2018multinomial} ,MAN\cite{chen2018multinomial}, CNN~\cite{cai2019multi} and BERT~\cite{cai2019multi}, among which CNN and BERT are the single-task learning methods) are listed in Table~\ref{table:keyword_methos_FDU} and Table~\ref{table:keyword_methos_amazon}. The column of “KCL-XXX” means we use XXX as the keyword extraction method, and the baseline is KCL-random, meaning that the order of domains is random. The table shows that using KeyBERT as the keyword extraction algorithm has the best effect on the FDU-MTL dataset, outperforming those without using the keyword extraction method by at least 1.13 points in average accuracy. At this time, the ranking of the FDU-MTL dataset is [Camera, Health, Kitchen, Software, MR, Apparel, Books, Magazine, Video, DVD, Music, Baby, Sports, IMDB, Toys, Electronics]. The table shows that using TextRank as the keyword extraction algorithm has the best effect on the Amazon review dataset, outperforming the random order by at least 1.79 points in average accuracy. At this time, the ranking of the Amazon review dataset is [DVD, Books, Electronics, Kitchen]. In addition, we can also see from this table that the performance of our KCL in each domain is also better than those single-task learning methods.

\subsection{Discussion: the number of keywords}
When calculating the obviousness of domain-specific features, different numbers of keywords may cause different ranking results of the sum of weights because there may exist such a situation that the sum of the weights of the top N1 keywords of a domain is larger than any other domains, but the sum of the weights of the first N2 (N2 \textgreater N1) keywords of that domain may be smaller than any other domains. Therefore, in this part, we conduct experiments to observe the influence of different numbers of keywords involved in weight calculation on the accuracy.

We test the effects of 30, 40, 50, 60, and 70 keywords. The results of accuracy are listed in Table~\ref{table:keyword_number_FDU} and Table~\ref{table:keyword_number_amazon}. Through the above experiments, we can see that when the number of keywords is 50, the effect is the best. 

\subsection{Implementation details}
We use bert-base-uncased as the shared feature extractor. Based on~\cite{chen2018multinomial}, we use a CNN consisting of an input layer, a single-layer convolutional layer, and a fully-connected layer for domain-specific feature extractors. In the convolutional layer, we use 3 sizes of convolution kernels. There are 200 convolution kernels for each size. After obtaining features of different scales through convolution kernels of different sizes, we then compress the output tensors of these 600 convolutions into 600 values through global max pooling and finally use these 600 values as the output of the convolution layer and input of the fully connected layer. 
More details of using CNN can be found in~\cite{soll2019evaluating}. 
The structures of the $Domain Discriminator$ and the $Text classifier$ are both very simple MLPs~\cite{liu2022we}.

\section{RELATED WORK}
 \textbf{Curriculum Learning}: Curriculum learning was proposed by Bengio~\cite{bengio2009curriculum}. The core of this strategy lies in the difficulty measurer of sample data and the data scheduling scheme of the training process. The sample difficulty measurer can be divided into automatic difficulty measurer and manual difficulty measurer. The automatic difficulty measurer refers to the measurement of the sample difficulty through the performance of the model itself, and the manual difficulty measurer refers to the grammatical and syntactic structure of the sample itself, such as the number of nouns and sentence length to measure the samples. After several years of development, curriculum learning strategies have shown excellent performance in response generation~\cite{shen2020cdl}, Contrastive Learning~\cite{ye2021efficient}.

\noindent\textbf{Multi-domain Text Classification}: Multi-domain text classification was initially proposed by Li S et al.~\cite{li2008multi}, and its goal is to improve the performance of a model in a specific domain by fusing data from multiple domains. Previously, the mainstream practice was to use the share-private structure to extract shared features and private features. That is, extracting the shared and domain-specific features. Later,  to prevent the confusion between shared features and domain-specific features, some teams began to use adversarial learning\cite{chen2018multinomial,wu2022co} to distinguish the two features. To make the model represent the text more accurately and train a better model, some teams applied the attention mechanism \cite{cai2019multi} and the pre-trained model\cite{luo2022mere} to the model.

\section{CONCLUSION}
\label{sec:majhead}

In this paper, we propose a network named KCL, which greatly improves the multi-domain text classification model in Amazon review datasets and FDU-MTL datasets. We first use the keyword extraction algorithm to calculate the weights of words in the corpus of each domain, take the N words with the highest weights as the keywords of the domain, and calculate the sum of the weights of the respective keywords in each domain. We regard the sum as the difficulty of extracting the domain-specific feature of a domain. The higher the sum of the weights is, the more obvious the domain-specific features of the domain are and the easier it is to extract. Therefore, in each step, the sample from the easier domain would enter the model earlier for training. Experimental results on amazon review datasets and FDU-MTL datasets show that our model reached the state-of-the-art of multi-domain text classification.

\clearpage

\bibliographystyle{IEEEbib}
\bibliography{strings,refs}

\end{document}